\documentclass[final]{cvpr}

\usepackage{times}
\usepackage{epsfig}
\usepackage{graphicx}
\usepackage{amsmath}
\usepackage{amssymb}
\usepackage{braket}
\usepackage{mathtools}
\usepackage{pifont}
\DeclarePairedDelimiter\abs{\lvert}{\rvert}%
\DeclarePairedDelimiter\norm{\lVert}{\rVert}%

\newcommand\blfootnote[1]{%
  \begingroup
  \renewcommand\thefootnote{}\footnote{#1}%
  \addtocounter{footnote}{-1}%
  \endgroup
}

\makeatletter
\let\oldabs\abs
\def\abs{\@ifstar{\oldabs}{\oldabs*}}
\let\oldnorm\norm
\def\norm{\@ifstar{\oldnorm}{\oldnorm*}}
\makeatother

\usepackage[pagebackref=true,breaklinks=true,colorlinks,bookmarks=false]{hyperref}

\def\cvprPaperID{8179} 
\def\confYear{CVPR 2021}

\begin{document}

\title{Shadow-Mapping for Unsupervised Neural Causal Discovery}

\author{Matthew J. Vowels\\
{\tt\small m.j.vowels@surrey.ac.uk}
\and
Necati Cihan Camgoz\\
{\tt\small n.camgoz@surrey.ac.uk}

\and
Richard Bowden\\
{\tt\small r.bowden@surrey.ac.uk}
\and
Centre for Vision, Speech and Signal Processing \\
University of Surrey\\
Guildford, UK\\
}

\maketitle

\begin{abstract}
An important goal across most scientific fields is the discovery of causal structures underling a set of observations. Unfortunately, causal discovery methods which are based on correlation or mutual information can often fail to identify causal links in systems which exhibit dynamic relationships. Such dynamic systems (including the famous coupled logistic map) exhibit `mirage' correlations which appear and disappear depending on the observation window. This means not only that correlation is not causation but, perhaps counter-intuitively, that causation may occur without correlation. In this paper we describe Neural Shadow-Mapping, a neural network based method which embeds high-dimensional video data into a low-dimensional shadow representation, for subsequent estimation of causal links. We demonstrate its performance at discovering causal links from video-representations of dynamic systems.
\end{abstract}

\blfootnote{Accepted to CVPR 2021 Causality in Vision Workshop.}

\section{Introduction}
Understanding causal structure is essential to the scientific endeavour \cite{Pearl2009}. Over recent decades, numerous methods have been proposed which seek to discover causal structure from data (for reviews, see \cite{Vowels2021DAGs,Runge2019, Heinze2018, Glymour2019}) but the task is inherently challenging. Numerous solutions may exist which are sufficient to explain the data, and causal links estimated using associational measures, such as correlation or mutual information, may fail in systems which exhibit complex state-based dependencies \cite{Sugihara2012}. These state dependent systems have been said to demonstrate \textit{mirage} correlations which appear and vanish over time. One such system is given by the well-known coupled logistic map difference equations \cite{May1976}:
\vspace{-1mm}
\begin{equation}
    \begin{split}
        X[n+1] = X[n](r_x - r_xX[n] - \beta_{xy}Y[n])\\
        Y[n+1] = Y[n](r_y - r_yY[n] - \beta_{yx}X[n])
    \end{split}
    \label{eq:logmap}
\end{equation}

\begin{figure}[t!]
\centering
\includegraphics[width=.65\linewidth]{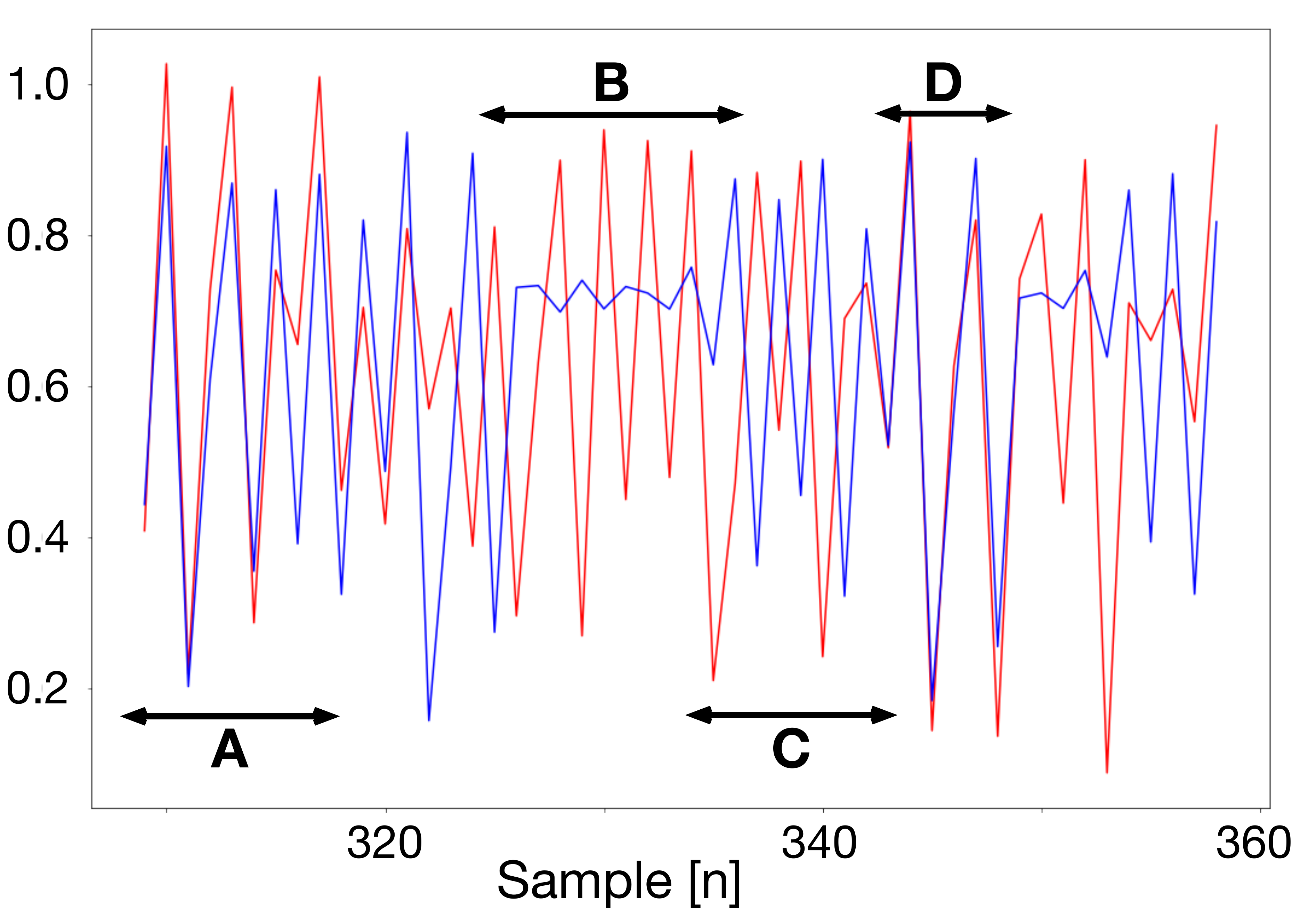}
\caption{Illustration of \textit{mirage} correlation for Eq. \ref{eq:logmap} with $r_x=3.8$, $r_y=3.8$, $\beta_{xy}=.02$, and $\beta_{yx}=.1$ (\textit{i.e.}, there exists bi-directional causality). Region \textbf{A} exhibits positive correlation, \textbf{B} low correlation, \textbf{C} negative correlation, and \textbf{D} returns to positive correlation. Example adapted from \cite{Sugihara2012}.}
\label{fig:mirage}
\end{figure}

Here, $X[n]$ and $Y[n]$ are two discrete-time varying quantities with parameters $r_x$ and $r_y$ and which causally influence each other via $\beta_{xy}$ and $\beta_{yx}$, respectively. Figure \ref{fig:mirage} demonstrates not only that correlation is not causation, but also that causation does not necessarily imply correlation, and thus a different approach is needed.

Dynamic systems, such as those described using Eq. \ref{eq:logmap}, occur frequently in nature \cite{May1976}, and it is therefore important that causal discovery methods can be applied. Unfortunately, it is understood that the most well-known and popular paradigm for modeling causal relations \textit{Granger Causality} does not perform well in such systems \cite{Sugihara2012}. This is because Granger causality assumes \textit{separability}, which refers to the independence of the variables in the absence of causal interactions. In dynamic systems, where the current state of a variable may be heavily determined by the past of another, separability is unlikely to hold.

\begin{figure}[t!]
\centering
\includegraphics[width=.6\linewidth]{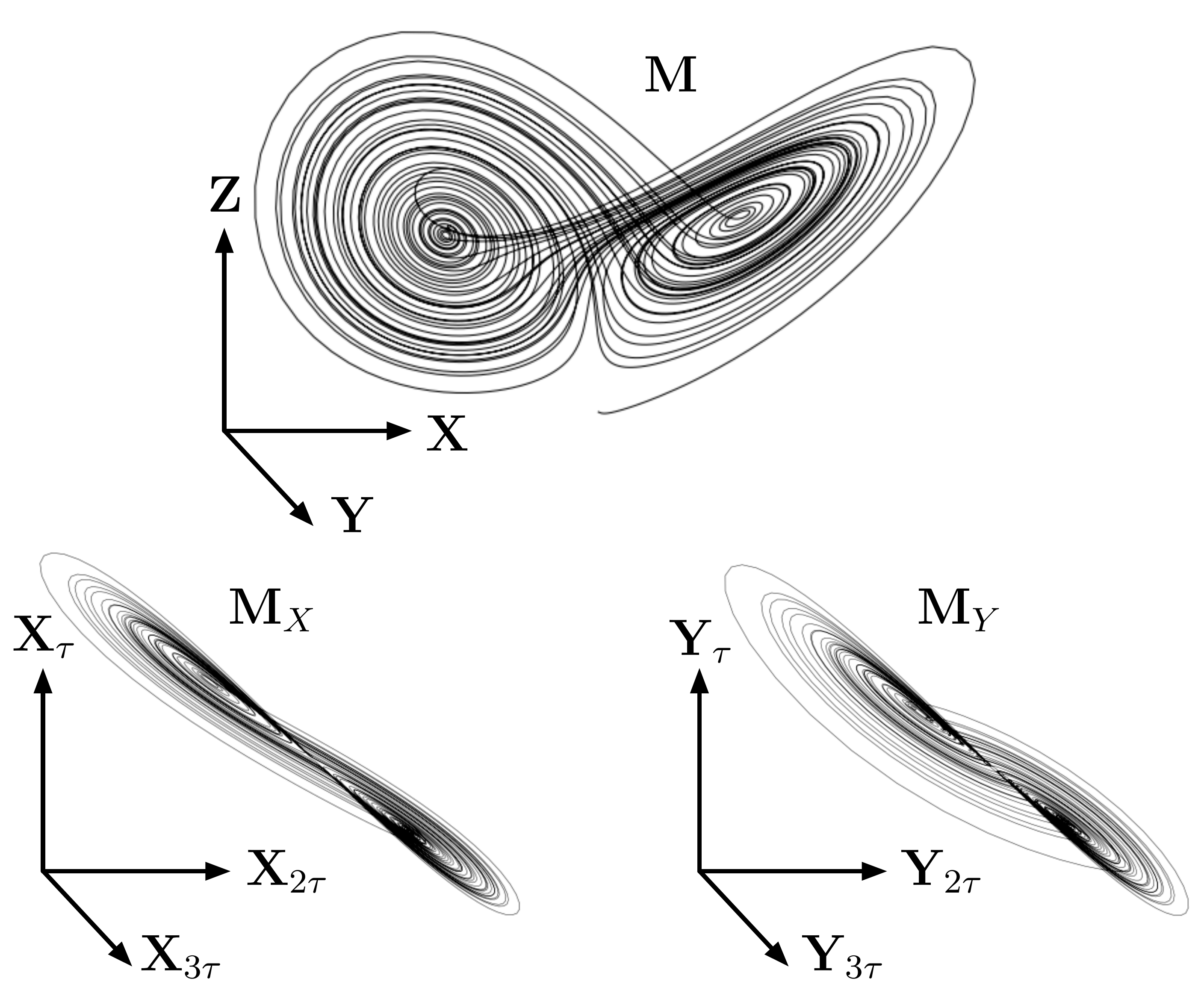}
\caption{A 3D Lorenz attractor $\mathbf{M}$ and corresponding shadow manifolds $\mathbf{M}_X$ and $\mathbf{M}_Y$ constructed using three lags (lag degree $\tau$).}
\label{fig:shadows}
\end{figure}

\textbf{Shadow Embeddings:} This failure of Granger Causality has motivated a family of causal discovery methods that operate on time-delayed coordinate embeddings. Takens \cite{Takens1981} showed that by concatenating time-delay versions of the time series observations, one can recover the dynamics of the full system even if only one, or a limited number, of observational variables are used. This time-delay embedding is known as a \textit{shadow manifold}. Specifically, assuming manifold $\mathbf{M}$ and $T$ time-based observations $\mathbf{X}: \mathbf{M} \rightarrow \mathbb{R}$, where $\mathbf{X} = \{x_1, x_2,... x_T\}$, the shadow manifold $\mathbf{M}_X$ is a Hankel matrix of delayed segments from $\mathbf{X}$:

\begin{equation}
   \mathbf{M}_X=\left[\begin{array}{cccc}x_1 & x_2 & \cdots & x_w \\ x_2 & x_3 & \cdots & x_{w+1} \\ \vdots & \vdots & \ddots & \vdots \\ x_p & x_{p+1} & \cdots & x_{p+w}\end{array}\right] 
\end{equation}

where $p$ is the number of lags, and $w$ is the segment/window size. An example of the Lorenz dynamic system being represented using two shadow manifolds is shown in Figure \ref{fig:shadows}. $\mathbf{M}_X$ and $\mathbf{M}_Y$ are diffeomorphic embeddings of the system $\mathbf{M}$.

\begin{figure}[b!]
\centering
\includegraphics[width=0.9\linewidth]{ 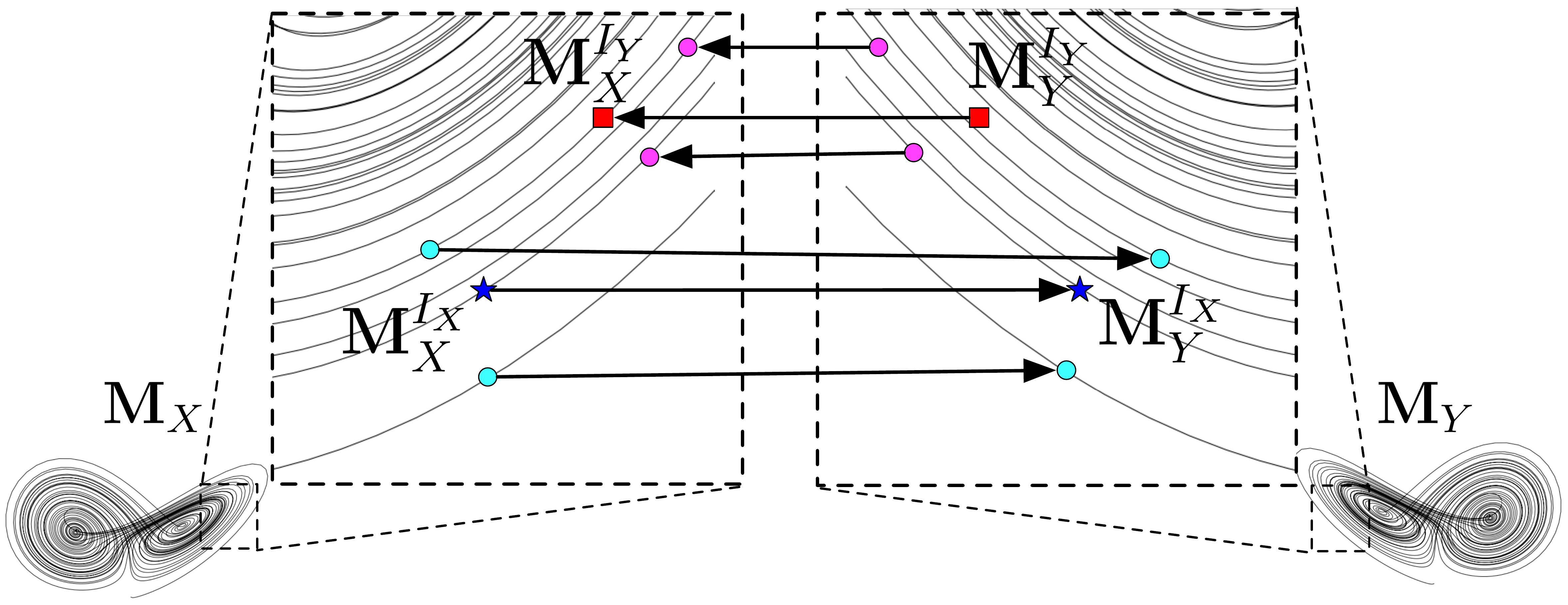}
\caption{Illustrating corresponding neighborhoods of points in the shadow manifolds (see text for details). }
\label{fig:crossmap}
\end{figure}

\textbf{Convergent Cross-Mapping (CCM):}  The embedding spaces can be used to identify causal links between the variables $\mathbf{X}$ and $\mathbf{Y}$ using CCM \cite{Sugihara2012}. Figure \ref{fig:crossmap} shows shadow manifolds $\mathbf{M}_X$ and $\mathbf{M}_Y$. Specifically, consider a particular time point in $\mathbf{M}_X$, illustrated with the star icon on the left hand side of the figure. The corresponding timepoint $t^*$ in $\mathbf{M}_Y$, indicated by the star in the right hand side of the figure. Additionally, there exists a collection of nearest neighbors illustrated with circles near to the point $\mathbf{M}_X^{t^*}$. The time indices $I_X$ of this set of nearest neighbors $\mathbf{M}_X^{I_X}$ may be far apart, despite their closeness in this shadow space. We can do the same thing in reverse for $M_Y$ by picking a time point $t^\Box$ (the square on the right hand side), and finding the time indices $I_Y$ for nearest neighbors $\mathbf{M}_Y^{I_Y}$. 
If the neighborhood of nearest neighbor indices $I_X$, derived from $\mathbf{M}_X$, induces a dense neighborhood $\mathbf{M}_Y^{I_X}$ then we may conclude that there exists a causal link $\mathbf{Y} \rightarrow \mathbf{X}$. This relationship can be tested in both directions (as illustrated in Figure \ref{fig:crossmap}) to test whether $\mathbf{X} \rightarrow \mathbf{Y}$. Furthermore, this process can be used to iteratively test more complex causal systems. For example, if the structure is confounding (\textit{i.e.},~a v-structure) such that $\mathbf{X} \leftarrow \mathbf{Z} \rightarrow \mathbf{Y}$, then the CCM test would establish this dependency structure.

\begin{figure}[t!]
\centering
\includegraphics[width=1\linewidth]{ 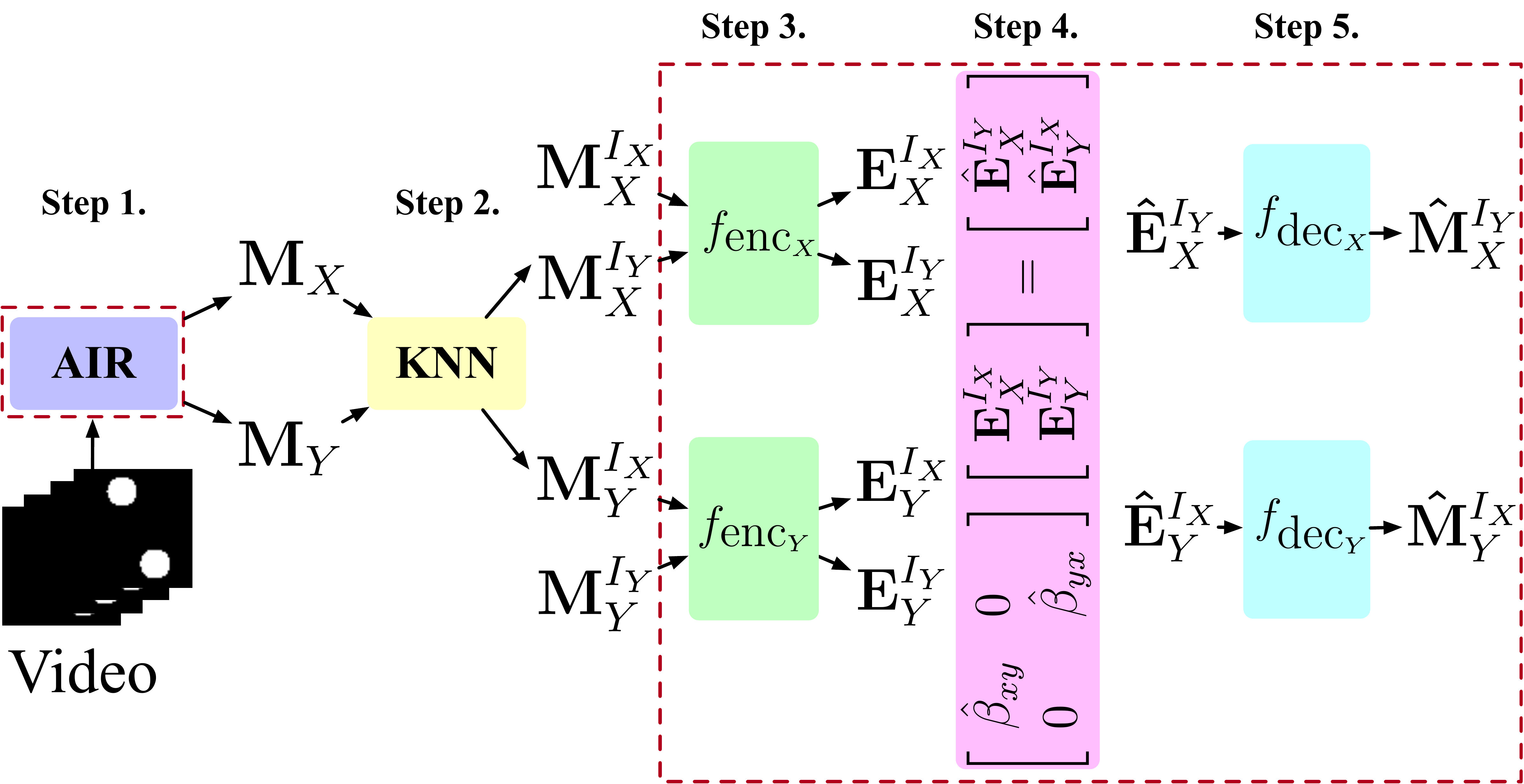}
\caption{Block diagram for NSM on a 2D example. Functions in red-dashed boxes are trained using gradient descent. Functions $f_{(.)}$ are fully connected neural networks.}
\label{fig:block}
\end{figure}

\textbf{Prior Work:}  A number of methods exist which leverage the principles behind CCM. Besides the original presentation of the method itself \cite{Sugihara2012}, it has been extended to identify lags \cite{Ye2015}, evaluated for its robustness to noise \cite{Monster2016}, improved using multivariate shadow embeddings \cite{McCracken2014}, and adapted for neural network integration \cite{Ma2014}. A related method involves the use of reservoir computing methods to improve upon the efficiency of CCM \cite{Huang2020b}.

More generally, methods fall into three predominant groups: score-based, constraint based, and asymmetry based. Score-based methods are learned by evaluating the fit of the observations to a proposed structure, constraint based methods test for statistical conditional independencies, and asymmetry based methods test for differences that arise when causality occurs in one direction compared with the other. Some of the most well-known methods for causal discovery are constraint based and include the PC algorithm \cite{Spirtes2000} and FCI \cite{Spirtes2000}, both of which test for conditional independencies in the data. Asymmetry based algorithms include LiNGAM \cite{Shimizu2006LiNGAM} and IGCI \cite{Janzing2012}, and score based methods include GES \cite{Chickering2002}, and neural network methods such as NOTEARS \cite{Zheng2018b}.

\textbf{Contributions: } Shadow embedding methods for discovering causal structure are not widely used in machine learning. In a recent review of over 100 causal discovery methods \cite{Vowels2021DAGs} only a small number of methods leveraged the principles behind CCM. The goals of this paper are therefore two-fold. (1) To introduce the principles behind CCM and dynamic systems causality to the machine learning community, with hopes for its wider adoption and exploration. (2) We present \textit{Neural Shadow-Mapping} (NSM), a method for causal discovery from video. As far as we are aware, NSM is the first application of the shadow-manifold causal discovery principles to image/video data.

\section{Method}
The block diagram for NSM is shown in Figure \ref{fig:block}. The process may be broken down into 5 steps. \textbf{ Step (1)}: Video frames are used to train a Pyro \cite{Bingham2019pyro} implementation of the unsupervised scene understanding method Attend, Infer, Repeat (AIR) \cite{Eslami2016}. This method provides frame-by-frame estimations of the object's positions. In the example used in the figure, there are two objects which each vary in their horizontal positions. There are therefore two system dimensions, and AIR provides time series for these dimensions $\mathbf{X}$ and $\mathbf{Y}$. \textbf{Step (2)}: Shadow embeddings $\mathbf{M}_X$ and $\mathbf{M}_Y$ are formed, and fed into a $k$-nearest neighbors algorithm \cite{Cover1967} to yield the neighborhoods indexed with the nearest neighbor indices $I$ from each embedding $\mathbf{M}_X$ and $\mathbf{M}_Y$, at each timepoint. For an example neighborhood at a single timepoint, we have \textit{e.g.}, $\mathbf{M}_X^{I_Y} \in \mathbb{R}^{p \times k} $, where $p$ is the number of lags used to form the shadow embedding, and $k$ is the number of nearest neighbors. We can form a batch over each indexed manifold by randomly selecting time points $t^*$ around which to form neighborhoods. \textbf{Step (3)}: $\mathbf{M}_X^{I_X}$ and $\mathbf{M}_X^{I_Y}$ are encoded using three fully-connected layers with non-linearity $f_{\mbox{enc}_X}$, whilst $\mathbf{M}_Y^{I_X}$ and $\mathbf{M}_Y^{I_Y}$ are fed through an equivalent fully-connected encoder $f_{\mbox{enc}_Y}$. Separate networks were used for each neighborhood-index/variable combination to maintain independence. \textbf{Step (4)}: The encoders yield new embeddings $\mathbf{E}^{(.)}_{(.)}$ of the indexed shadow embeddings. The motivation for the encoder is yield a lower-dimensional representation ameanable to interpretable linear regression: $\mathbf{A}\mathbf{\bar{E}} = \mathbf{\hat{E}}$ where $\mathbf{A}$ is a learnable, diagonal weight matrix intended to discover the causal links $\beta_{xy}$ and $\beta_{yx}$. $\mathbf{\bar{E}}$ is a stacked matrix of row vectors $\mathbf{E}_X^{I_X}$ $\mathbf{E}_Y^{I_Y}$, that we use to predict $\mathbf{\hat{E}}$, which constitutes $\mathbf{E}_X^{I_Y}$ and $\mathbf{E}_Y^{I_X}$, respectively. This step (4) therefore represents a modified version of the CCM element of the process, because traditional cross-mapping would use e.g. $\mathbf{E}_X^{I_X}$ to predict $\mathbf{E}_Y^{I_X}$ - the superscript indices $I_X$ would be used to index $Y$. In practice we found that mapping `within-variable' using different indices yields better results. The resulting $\hat\beta$ coefficients were biased due to the inherent correlation for within-variable mapping, but this can be accounted for in the surrogate testing process. An $l_2$ loss between $\mathbf{\hat{E}}$ and $[\mathbf{E}_X^{I_Y},\mathbf{E}_Y^{I_X}]$ is used to optimize the non-zero parameters in $\mathbf{A}$. \textbf{Step (5)}: The final step involves reconstruction of the indexed manifolds $\mathbf{M}_X^{I_Y}$ and $\mathbf{M}_Y^{I_X}$ from the estimated $\mathbf{E}_X^{I_Y}$ and $\mathbf{E}_Y^{I_X}$ via fully-connected decoders $f_{\mbox{dec}_X}$ and $f_{\mbox{dec}_Y}$, respectively. The error on the reconstructions is measured using the $l_2$ loss which is also used to drive learning via backpropagation for steps (3)-(5).

\begin{figure}[b!]
\centering
\includegraphics[width=.55\linewidth]{ 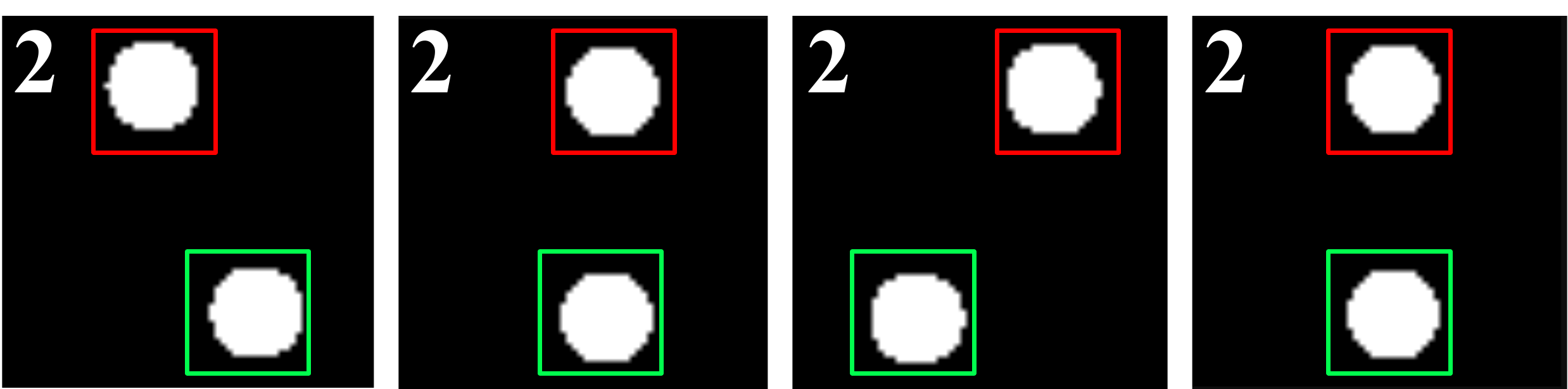}
\caption{Four timesteps from the two-variable video, overlaid bounding boxes from AIR \cite{Eslami2016}.}
\label{fig:vidsamples}
\end{figure}

\begin{table}[]
\footnotesize
\centering
\begin{tabular}{lccc} \hline
Video Graph  &   $p$-val. ($\hat\beta_{XY}$)&  $p$-val.  ($\hat\beta_{YX}$) & Identified?\\ \hline
$X \; \; \; \; \; \;Y$ & 0.61 &  0.24   & \checkmark\\
$X\rightarrow Y$      &  0.24 &  1.88e-6  & \checkmark  \\
 $X\leftarrow Y$       &   3.31e-8 &  0.45 & \checkmark \\
 $X\leftrightarrow Y$       & 3.16e-19 &  1.66e-12  & \checkmark \\ \hline Time-Series Graph
 &  & Threshold & Identified?\\ \hline
 $X \rightarrow Y \; \; \; \; \; Z$ &  & 0.25    & \checkmark \\ 
 $X \leftrightarrow Y \; \; \; \; \; Z$ &  & 0.25    & \checkmark \\ 
 $X \rightarrow Z\rightarrow Y\rightarrow X$ &  & 0.25    & \checkmark \\
 $X \leftrightarrow Z\rightarrow Y\rightarrow X$ &  & 0.25    & \checkmark \\
  $Y \leftarrow X \rightarrow Z$ &  & 0.25    & \checkmark \\
  $X \rightarrow Z \leftarrow Y$ &  & 0.25    & \checkmark \\ 
  $X \leftrightarrow Z \leftarrow Y$ &  & 0.25    & \checkmark \\
    $X \leftrightarrow Z \rightarrow Y$ &  & 0.25    & \checkmark \\ 
    $X \leftrightarrow Z \leftrightarrow Y$ &  & 0.25    & \checkmark \\ 
  \hline \hline
\end{tabular}
\caption{Upper: $p$-values from a one sided, 2-sample KS test between estimated path coefficients on video embeddings of bivariate data with the structure represented by corresponding graphs, and the IAAFT surrogates of those data ($p > 0.01$ means no effect). Lower: the lower portion is for trivariate time-series data. }
\label{tab:one}
\end{table}

\section{Experiments}
In order to demonstrate this method, two synthetic datasets were created. The first is a video dataset, whereby the horizontal positions of two objects varied according to the coupled logistic map in Equation \ref{eq:logmap}. Four time steps are given as example in Figure \ref{fig:vidsamples}. To explore the 3-variable case, we also create tri-variate time series data according to the equation below:
\vspace{-2mm}
\begin{equation}
    \begin{split}
        X_i[n+1] = X_i[n](r_i - r_iX_i[n] - \beta_{ij}X_j[n] - \beta_{ik}X_k[n])\\
    \end{split}
    \label{eq:logmap}
\end{equation}


for $i,j,k=\{1,2,3\}$.\footnote{In Table \ref{tab:one}, $X_{\{1,2,3\}}$ are equivalent to $X,Y,Z$.} We set all $r_{(.)}=3.9$ and all non-zero $\beta_{(.,.)}=0.25$. For both datasets, data are generated using random $\sim U[0,1]$ initial conditions, and the learned parameter matrix $\mathbf{A}$ is amortized over all these generations. The training process is repeated to acquire distributions over the estimated parameters $\hat{\boldsymbol{\beta}}$ in $\mathbf{A}$, which are then inspected to discover causal connections. The discovery process itself involves testing for significant increase above a non-causal baseline. This baseline is established by creating time series surrogates based on the Iterative Amplitude Adjusted Fourier Transform (IAAFT) method \cite{Lucio2012, Schreiber1996}. IAAFT creates surrogates which match the original data in terms of their power spectra and therefore also their autocorrelation.\footnote{We also validated these surrogates by creating non-causal system data (\textit{i.e.}, by setting $\beta_{xy} = \beta_{yx} = 0$) and comparing the results with the IAAFT results.} The set of resulting baseline parameters can be used as a `null' distribution and tested against the parameters of the original data using a one-sided, two-sample Kolmogorov Smirnoff (KS) test. If the KS test statistic has a $p$-value below the false-positive threshold $\alpha$, then one may infer statistical significance. We found that the KS test was not required for the trivariate time-series data, and that a simple fixed threshold of 0.25 could be used to infer the presence of an effect. For the experiments, we set the number of runs $N_r=100$, $\alpha=0.01$, $p=10$, $w=790$, $k=10$, time series length $=\{20, 1000\}$, embedding dimension for $\mathbf{E}=6$, batch size $=20$, number of training iterations $3e5$, learning rate $3e-4$, and use an Adam \cite{adam} optimizer. No hyperparameter tuning was performed.

The results for the two variable and three variable datasets are shown in Table \ref{tab:one} where `identified?' indicates whether a link was discovered (but not the magnitude of the link). NSM successfully recovered the true graph in all cases. 



\section{Discussion and Limitations}
NSM was able to discover causal links from video and time-series data generated from dynamic systems. This is a notable result, particularly because visually identifying causality from the video is non-trivial. As is the case for traditional CCM, the discovery of causal links is sensitive to measurement noise and stochasticity. Furthermore, the disadvantages associated with nearest-neighbor methods, include the need for longer time series. Future work should identify ways to improve its robustness to noise, possibly by using existing techniques, and to operate with shorter time series \cite{Monster2016}. The aim should then be to test the generalization of the method in discovering causal links in more challenging problems such as those involving human interaction or traffic data.

{\small
\bibliographystyle{ieee_fullname}
\bibliography{cvpr}
}

\end{document}